# NONLINEAR CHANNEL ESTIMATION FOR OFDM SYSTEM BY COMPLEX LS-SVM UNDER HIGH MOBILITY CONDITIONS

Anis Charrada[1] and Abdelaziz Samet[2]

[1 and 2] CSE Research Unit, Tunisia Polytechnic School, Carthage University,
Tunis, Tunisia .
[1]`anis.charrada@gmail.com`, [2]`abdelaziz.samet@ept.rnu.tn`

## ABSTRACT

*A nonlinear channel estimator using complex Least Square Support Vector Machines (LS-SVM) is proposed for pilot-aided OFDM system and applied to Long Term Evolution (LTE) downlink under high mobility conditions. The estimation algorithm makes use of the reference signals to estimate the total frequency response of the highly selective multipath channel in the presence of non-Gaussian impulse noise interfering with pilot signals. Thus, the algorithm maps trained data into a high dimensional feature space and uses the structural risk minimization (SRM) principle to carry out the regression estimation for the frequency response function of the highly selective channel. The simulations show the effectiveness of the proposed method which has good performance and high precision to track the variations of the fading channels compared to the conventional LS method and it is robust at high speed mobility.*

## *Keywords*

*Complex LS-SVM, Mercer's kernel, nonlinear channel estimation, impulse noise, OFDM, LTE.*

## 1. INTRODUCTION

Channel estimation in wireless OFDM systems is an active research area, especially in the case of frequency selective time varying multipath fading channels. Several estimation algorithms have been developed, such that LS [1], MMSE [2] and estimation with decision feedback [3]. Channel estimation by neural network is also described in [4]. However, in a practical environment where non-Gaussian impulse noise can be present, the classical estimation methods may not be effective for this impulse noise.

The use of Support Vector Machines (SVMs) has already been proposed to solve a variety of signal processing and digital communications problems. Signal equalization and detection for multicarrier MC-CDMA system is presented in [5]. Also, adaptive multiuser detector for direct sequence CDMA signals in multipath channels is developed in [6]. In all these applications, SVM methods outperform classical approaches due to its improved generalization capabilities.

Here, a proposed SVM robust version for nonlinear channel estimation in the presence of non-Gaussian impulse noise that is specifically adapted to pilot-aided OFDM structure is presented. In fact, impulses of short duration are unpredictable and contain spectral components on all subchannels which impact the decision of the transmitted symbols on all subcarriers.

The channel estimation algorithm is based on the nonlinear least square support vector machines (LS-SVM) method in order to improve communication efficiency and quality of OFDM systems. The principle of the proposed nonlinear LS-SVM algorithm is to exploit the information provided by the reference signal to estimate the channel frequency response. In highly selective multipath fading channel, where complicated nonlinearities can be present, the estimation precision can be lowed by using linear method. So, we adapt the nonlinear LS-SVM algorithm which transforms the nonlinear estimation in low dimensional space into the linear estimation in high dimensional space, so it improves the estimation precision.

In this contribution, the proposed nonlinear complex LS-SVM technique is applied to LTE downlink highly selective channel using pilot symbols. For the purpose of comparison with conventional LS algorithm, we develop the nonlinear LS-SVM algorithm in terms of the RBF kernel. Simulation section illustrates the advantage of this algorithm over LS algorithm in high mobility environment. The nonlinear complex LS-SVM method shows good results under high mobility conditions due to its improved generalization ability.

The scheme of the paper is as follows. Section 2 briefly introduces the OFDM system model. We present the formulation of the proposed nonlinear complex LS-SVM channel estimation method in section 3. Section 4 presents the simulation results when comparing with LS standard algorithm. Finally, in section 5, conclusions are drawn.

## 2. SYSTEM MODEL

The OFDM system model consists firstly of mapping binary data streams into complex symbols by means of QAM modulation. Then data are transmitted in frames by means of serial-to-parallel conversion. Some pilot symbols are inserted into each data frame which is modulated to subcarriers through IDFT. These pilot symbols are inserted for channel estimation purposes. The IDFT is used to transform the data sequence $X(k)$ into time domain signal as follow:

$$x(n) = IDFT_N\{X(k)\} = \sum_{k=0}^{N-1} X(k) e^{j\frac{2\pi}{N}kn}, \qquad n = 0, \cdots, N-1 \qquad (1)$$

One guard interval is inserted between every two OFDM symbols in order to eliminate inter-symbol interference (ISI). This guard time includes the cyclically extended part of the OFDM symbol in order to preserve orthogonality and eliminate inter-carrier interference (ICI). It is well known that if the channel impulse response has a maximum of $L$ resolvable paths, then the GI must be at least equal to $L$ [7].

Thus, for the OFDM system comprising $N$ subcarriers which occupy a bandwidth $B$, each OFDM symbol is transmitted in time $T$ and includes a cyclic prefix of duration $T_{cp}$. Therefore, the duration of each OFDM symbol is $T_u = T - T_{cp}$. Every two adjacent subcarriers are spaced by $\delta f = 1/T_u$. The output signal of the OFDM system is converted into serial signal by parallel to serial converter. A complex white Gaussian noise process $N(0, \sigma_w^2)$ with power spectral density $N_0/2$ is added through a frequency selective time varying multipath fading channel.

In a practical environment, impulse noise can be present, and then the channel becomes nonlinear with non Gaussian impulse noise. The impulse noise can significantly influence the performance of the OFDM communication system for many reasons. First, the time of the arrival of an impulse is unpredictable and shapes of the impulses are not known and they vary considerably. Moreover, impulses usually have very high amplitude, and thus high energy, which can be much greater than the energy of the useful signal [8].

The impulse noise is modeled as a Bernoulli-Gaussian process and it was generated with the Bernoulli-Gaussian process function $i(n) = v(n)\lambda(n)$ where $v(n)$ is a random process with Gaussian distribution and power $\sigma_{BG}^2$, and where $\lambda(n)$ is a random process with probability [9]

$$P_r(\lambda(n)) = \begin{cases} p & \lambda = 1 \\ 1-p, & \lambda = 0. \end{cases} \qquad (2)$$

At the receiver, and after removing guard time, the discrete-time baseband OFDM signal for the system including impulse noise is

$$y(n) = \sum_{k=0}^{N-1} X(k)H(k)e^{j\frac{2\pi}{N}kn} + w(n) + i(n), \qquad n = 0, \cdots, N-1 \qquad (3)$$

where $y(n)$ are time domain samples and $H(k) = DFT_N\{h(n)\}$ is the channel's frequency response at the $k^{th}$ frequency. The sum of both terms of the AWGN noise and impulse noise constitute the total noise given by $z(n) = w(n) + i(n)$.

Let $\Omega_P$ the subset of $N_P$ pilot subcarriers and $\Delta P$ the pilot interval in frequency domain. Over this subset, channel's frequency response can be estimated, and then interpolated over other subcarriers $(N - N_P)$. These remaining subchannels are interpolated by the nonlinear complex LS-SVM algorithm. The OFDM system can be expressed as

$$y(n) = y^P(n) + y^D(n) + z(n)$$

$$= \sum_{k \in \{\Omega_P\}} X^P(k)H(k)e^{j\frac{2\pi}{N}kn} + \sum_{k \notin \{\Omega_P\}} X^D(k)H(k)e^{j\frac{2\pi}{N}kn} + z(n) \qquad (4)$$

where $X^P(k)$ and $X^D(k)$ are complex pilot and data symbol respectively, transmitted at the $k^{th}$ subcarrier. Note that, pilot insertion in the subcarriers of every OFDM symbol must satisfy the demand of the sampling theory and uniform distribution [10].

After DFT transformation, $y(n)$ becomes

$$Y(k) = DFT_N\{y(n)\} = \frac{1}{N} \sum_{n=0}^{N-1} y(n) e^{-j\frac{2\pi}{N}kn}, \quad k = 0, \cdots, N-1 \qquad (5)$$

Assuming that ISI are eliminated, therefore

$$Y(k) = X(k)H(k) + W(k) + I(k) = X(k)H(k) + e(k), \quad k = 0, \cdots, N-1 \qquad (6)$$

where $e(k)$ represents the sum of the AWGN noise $W(k)$ and impulse noise $I(k)$ in the frequency domain, respectively.

(6) may be presented in matrix notation

$$Y = XFh + W + I = XH + e \qquad (7)$$

where

$$\begin{aligned} X &= diag(X(0), X(1), \cdots, X(N-1)) \\ Y &= [Y(0), \cdots, Y(N-1)]^T \\ W &= [W(0), \cdots, W(N-1)]^T \\ I &= [I(0), \cdots, I(N-1)]^T \\ H &= [H(0), \cdots, H(N-1)]^T \\ e &= [e(0), \cdots, e(N-1)]^T \end{aligned}$$

$$F = \begin{bmatrix} W_N^{00} & \cdots & W_N^{0(N-1)} \\ \vdots & \ddots & \vdots \\ W_N^{(N-1)0} & \cdots & W_N^{(N-1)(N-1)} \end{bmatrix}$$

and
$$W_N^{i,k} = \left(\frac{1}{\sqrt{N}}\right) exp^{-j2\pi\left(\frac{ik}{N}\right)}. \qquad (8)$$

International Journal of Wireless & Mobile Networks (IJWMN) Vol. 3, No. 4, August 2011

## 3. NONLINEAR COMPLEX LS-SVM ESTIMATOR

First, let the OFDM frame contains $Ns$ OFDM symbols which every symbol includes $N$ subcarriers. As every OFDM symbol has $N_P$ uniformly distributed pilot symbols, the transmitting pilot symbols are $X^P = diag(X(s, m\,\Delta P)), m = 0,1,\cdots N_P - 1$, where $s$ and $m$ are labels in time domain and frequency domain respectively.

The proposed channel estimation method is based on nonlinear complex LS-SVM algorithm which has two separate phases: training phase and estimation phase. In training phase, we estimate first the subchannels pilot symbols according to LS criterion to strike $min\,[(Y^P - X^P Fh)(Y^P - X^P Fh)^H]$ [11], as

$$\hat{H}^P = X^{P\,-1} Y^P \tag{9}$$

where $Y^P = Y(s, m\,\Delta P)$ and $\hat{H}^P = \hat{H}(s, m\,\Delta P)$ are the received pilot symbols and the estimated frequency responses for the $s^{th}$ OFDM symbol at pilot positions $m\,\Delta P$, respectively. Then, in the estimation phase and by the interpolation mechanism, frequency responses of data subchannels can be determined. Therefore, frequency responses of all the OFDM subcarriers are

$$\hat{H}(s,q) = f\left(\hat{H}^P(s, m\,\Delta P)\right) \tag{10}$$

where $q = 0,\cdots, N-1$, and $f(\cdot)$ is the interpolating function, which is determined by the nonlinear complex LS-SVM approach.

In high mobility environments, where the fading channels present very complicated nonlinearities especially in deep fading case, the linear approaches cannot achieve high estimation precision. Therefore, we adapt here a nonlinear complex LS-SVM technique since SVM is superior in solving nonlinear, small samples and high dimensional pattern recognition [10]. Therefore, we map the input vectors to a higher dimensional feature space $\mathcal{H}$ (possibly infinity) by means of nonlinear transformation $\boldsymbol{\varphi}$. Thus, the regularization term is referred to the regression vector in the RKHS. The following regression function is then

$$\hat{H}(m\,\Delta P) = \boldsymbol{w}^T \boldsymbol{\varphi}(m\,\Delta P) + b + e_m, \quad m = 0,\cdots N_P - 1 \tag{11}$$

where $\boldsymbol{w}$ is the weight vector, $b$ is the bias term well known in the SVM literature and residuals $\{e_m\}$ account for the effect of both approximation errors and noise. In the SVM framework, the optimality criterion is a regularized and constrained version of the regularized LS criterion. In general, SVM algorithms minimize a regularized cost function of the residuals, usually the Vapnik's $\varepsilon - insensitivity$ cost function [9]. A robust cost function is introduced to improve the performance of the estimation algorithm which is $\varepsilon$ -Huber robust cost function, given by [12]

$$\mathcal{L}^\varepsilon(e_m) = \begin{cases} 0, & |e_m| \leq \varepsilon \\ \dfrac{1}{2\gamma}(|e_m| - \varepsilon)^2, & \varepsilon \leq |e_m| \leq e_C \\ C(|e_m| - \varepsilon) - \dfrac{1}{2}\gamma C^2, & e_C \leq |e_m| \end{cases} \tag{12}$$

where $e_C = \varepsilon + \gamma C$, $\varepsilon$ is the insensitive parameter which is positive scalar that represents the insensitivity to a low noise level, parameters $\gamma$ and $C$ control essentially the trade-off between the regularization and the losses, and represent the relevance of the residuals that are in the linear or in the quadratic cost zone, respectively. The cost function is linear for errors above $e_C$, and quadratic for errors between $\varepsilon$ and $e_C$. Note that, errors lower than $\varepsilon$ are ignored in the $\varepsilon - insensitivite$ zone. On the other hand, the quadratic cost zone uses the $L_2 - norm$ of errors, which is appropriate for Gaussian noise, and the linear cost zone limits the effect of sub-Gaussian noise [13]. Therefore, the $\varepsilon$ -Huber robust cost function can be adapted to different kinds of noise.

Since $e_m$ is complex, let $\mathcal{L}^\varepsilon(e_m) = \mathcal{L}^\varepsilon(\mathcal{R}(e_m)) + \mathcal{L}^\varepsilon(\mathfrak{I}(e_m))$, where $\mathcal{R}(\cdot)$ and $\mathfrak{I}(\cdot)$ represent real and imaginary parts, respectively.

Now, we can state the primal problem as minimizing

$$\frac{1}{2}\|\boldsymbol{w}\|^2 + \frac{1}{2\gamma}\sum_{m\in I_1}(\xi_m + \xi_m^*)^2 + C\sum_{m\in I_2}(\xi_m + \xi_m^*) + \frac{1}{2\gamma}\sum_{m\in I_3}(\zeta_m + \zeta_m^*)^2$$
$$+ C\sum_{m\in I_4}(\zeta_m + \zeta_m^*) - \frac{1}{2}\sum_{m\in I_2, I_4}\gamma C^2 \qquad (13)$$

constrained to

$$\mathcal{R}(\widehat{H}(m\,\Delta P) - \boldsymbol{w}^T\boldsymbol{\varphi}(m\,\Delta P) - b) \leq \varepsilon + \xi_m$$

$$\mathfrak{I}(\widehat{H}(m\,\Delta P) - \boldsymbol{w}^T\boldsymbol{\varphi}(m\,\Delta P) - b) \leq \varepsilon + \zeta_m$$

$$\mathcal{R}(-\widehat{H}(m\,\Delta P) + \boldsymbol{w}^T\boldsymbol{\varphi}(m\,\Delta P) + b) \leq \varepsilon + \xi_m^*$$

$$\mathfrak{I}(-\widehat{H}(m\,\Delta P) + \boldsymbol{w}^T\boldsymbol{\varphi}(m\,\Delta P) + b) \leq \varepsilon + \zeta_m^*$$

$$\xi_m^{(*)}, \zeta_m^{(*)} \geq 0 \qquad (14)$$

for $m = 0, \cdots, N_P - 1$, where $\xi_m$ and $\xi_m^*$ are slack variables which stand for positive and negative errors in the real part, respectively. $\zeta_m$ and $\zeta_m^*$ are the errors for the imaginary parts. $I_1, I_2, I_3$ and $I_4$ are the set of samples for which:

$I_1$ : real part of the residuals are in the quadratic zone;

$I_2$ : : real part of the residuals are in the linear zone;

$I_3$ : : imaginary part of the residuals are in the quadratic zone;

$I_4$ : : imaginary part of the residuals are in the linear zone.

To transform the minimization of the primal functional (13) subject to constraints in (14), into the optimization of the dual functional, we must first introduce the constraints into the primal functional. Thus, the primal dual functional is as follow:

$$L_{Pd} = \frac{1}{2}\|\boldsymbol{w}\|^2 + \frac{1}{2\gamma}\sum_{m\in I_1}(\xi_m + \xi_m^*)^2 + C\sum_{m\in I_2}(\xi_m + \xi_m^*) + \frac{1}{2\gamma}\sum_{m\in I_3}(\zeta_m + \zeta_m^*)^2$$
$$+ C\sum_{m\in I_4}(\zeta_m + \zeta_m^*) - \frac{1}{2}\sum_{m\in I_2, I_4}\gamma C^2 - \sum_{m=0}^{N_P-1}(\beta_m\xi_m + \beta_m^*\xi_m^*) - \sum_{m=0}^{N_P-1}(\lambda_m\zeta_m + \lambda_m^*\zeta_m^*)$$
$$+ \sum_{m=0}^{N_P-1}\alpha_{\mathcal{R},m}\left[\mathcal{R}(\widehat{H}(m\,\Delta P) - \boldsymbol{w}^T\boldsymbol{\varphi}(m\,\Delta P) - b) - \varepsilon - \xi_m\right]$$

4International Journal of Wireless & Mobile Networks (IJWMN) Vol. 3, No. 4, August 2011$$+ \sum_{m=0}^{N_P-1} \alpha_{I,m} \left[ \Im\left(\widehat{H}(m\,\Delta P) - \mathbf{w}^T \boldsymbol{\varphi}(m\,\Delta P) - b\right) - j\varepsilon - j\zeta_m \right]$$

$$+ \sum_{m=0}^{N_P-1} \alpha_{\mathcal{R},m}^* \left[ \mathcal{R}\left(-\widehat{H}(m\,\Delta P) + \mathbf{w}^T \boldsymbol{\varphi}(m\,\Delta P) + b\right) - \varepsilon - \xi_m^* \right]$$

$$+ \sum_{m=0}^{N_P-1} \alpha_{I,m}^* \left[ \Im\left(-\widehat{H}(m\,\Delta P) + \mathbf{w}^T \boldsymbol{\varphi}(m\,\Delta P) + b\right) - j\varepsilon - j\zeta_m^* \right] \tag{15}$$

with the Lagrange multipliers (or dual variables) constrained to $\alpha_{\mathcal{R},m}, \alpha_{I,m}, \beta_m, \lambda_m, \alpha_{\mathcal{R},m}^*, \alpha_{I,m}^*, \beta_m^*, \lambda_m^* \geq 0$ and $\xi_m, \zeta_m, \xi_m^*, \zeta_m^* \geq 0$.

According to Karush-Kuhn-Tucker (KKT) conditions [12]

$$\beta_m \xi_m = 0,\ \beta_m^* \xi_m^* = 0 \text{ and } \lambda_m \zeta_m = 0,\ \lambda_m^* \zeta_m^* = 0. \tag{16}$$

Then, by making zero the primal-dual functional gradient with respect to $\varpi_i$, we obtain an optimal solution for the weights

$$\mathbf{w} = \sum_{m=0}^{N_P-1} \psi_m \boldsymbol{\varphi}(m\,\Delta P) = \sum_{m=0}^{N_P-1} \psi_m \boldsymbol{\varphi}(P_m) \tag{17}$$

where $\psi_m = (\alpha_{\mathcal{R},m} - \alpha_{\mathcal{R},m}^*) + j(\alpha_{I,m} - \alpha_{I,m}^*)$ with $\alpha_{\mathcal{R},m}, \alpha_{\mathcal{R},m}^*, \alpha_{I,m}, \alpha_{I,m}^*$ are the Lagrange multipliers for real and imaginary part of the residuals and $P_m = (m\,\Delta P),\ m = 0,\cdots,N_P - 1$ are the pilot positions.

We define the Gram matrix as

$$\mathbf{G}(u,v) = <\boldsymbol{\varphi}(P_u), \boldsymbol{\varphi}(P_v)> = K(P_u, P_v) \tag{18}$$

where $K(P_u, P_v)$ is a Mercer's kernel which represent in this contribution the RBF kernel matrix which allows obviating the explicit knowledge of the nonlinear mapping $\boldsymbol{\varphi}(\cdot)$. A compact form of the functional problem can be stated in matrix format by placing optimal solution $\mathbf{w}$ into the primal dual functional and grouping terms. Then, the dual problem consists of maximizing

$$-\frac{1}{2}\boldsymbol{\psi}^H(\mathbf{G} + \gamma\mathbf{I})\boldsymbol{\psi} + \mathcal{R}(\boldsymbol{\psi}^H Y^P) - (\boldsymbol{\alpha}_{\mathcal{R}} + \boldsymbol{\alpha}_{\mathcal{R}}^* + \boldsymbol{\alpha}_I + \boldsymbol{\alpha}_I^*)\mathbf{1}\varepsilon \tag{19}$$

constrained to $0 \leq \alpha_{\mathcal{R},m}, \alpha_{\mathcal{R},m}^*, \alpha_{I,m}, \alpha_{I,m}^* \leq C$, where $\boldsymbol{\psi} = [\psi_0,\cdots,\psi_{N_P-1}]^T$; $\mathbf{I}$ and $\mathbf{1}$ are the identity matrix and the all-ones column vector, respectively; $\boldsymbol{\alpha}_{\mathcal{R}}$ is the vector which contains the corresponding dual variables, with the other subsets being similarly represented. The weight vector can be obtained by optimizing (19) with respect to $\alpha_{\mathcal{R},m}, \alpha_{\mathcal{R},m}^*, \alpha_{I,m}, \alpha_{I,m}^*$ and then substituting into (17).

Therefore, and after training phase, frequency responses at all subcarriers in each OFDM symbol can be obtained by SVM interpolation

$$\widehat{H}(k) = \sum_{m=0}^{N_P-1} \psi_m K(P_m, k) + b \tag{20}$$

for $k = 1,\cdots,N$. Note that, the obtained subset of Lagrange multipliers which are nonzero will provide with a sparse solution. As usual in the SVM framework, the free parameter of the kernel and the free parameters of the cost function have to be fixed by some a priori knowledge of the problem, or by using some validation set of observations [9].

180

## 4. SIMULATION RESULTS

We consider the channel impulse response of the frequency-selective fading channel model which can be written as

$$h(\tau, t) = \sum_{l=0}^{L-1} h_l(t)\, \delta(t - \tau_l) \quad (21)$$

where $h_l(t)$ is the impulse response representing the complex attenuation of the $l^{th}$ path, $\tau_l$ is the random delay of the $l^{th}$ path and $L$ is the number of multipath replicas. The specification parameters of an extended vehicular A model (EVA) for downlink LTE system with the excess tap delay and the relative power for each path of the channel are shown in table 1. These parameters are defined by 3GPP standard [14].

Table 1. Extended Vehicular A model (EVA) [14].

| Excess tap delay [ns] | Relative power [dB] |
|---|---|
| 0 | 0.0 |
| 30 | -1.5 |
| 150 | -1.4 |
| 310 | -3.6 |
| 370 | -0.6 |
| 710 | -9.1 |
| 1090 | -7.0 |
| 1730 | -12.0 |
| 2510 | -16.9 |

In order to demonstrate the effectiveness of our proposed technique and evaluate the performance, two objective criteria, the signal-to-noise ratio (SNR) and signal-to-impulse ratio (SIR) are used. The SNR and SIR are given by [9]

$$SNR_{dB} = 10 log_{10} \left( \frac{E\{|y(n) - w(n) - i(n)|^2\}}{\sigma_w^2} \right) \quad (22)$$

and

$$SIR_{dB} = 10 log_{10} \left( \frac{E\{|y(n) - w(n) - i(n)|^2\}}{\sigma_{BG}^2} \right) \quad (23)$$

Then, we simulate the OFDM downlink LTE system with parameters presented in table 2. The nonlinear complex LS-SVM estimate a number of OFDM symbols in the range of 140 symbols, corresponding to one radio frame LTE. Note that, the LTE radio frame duration is 10 ms [15], which is divided into 10 subframes. Each subframe is further divided into two slots, each of 0.5 ms duration.

For the purpose of evaluation the performance of the nonlinear complex LS-SVM algorithm under high mobility conditions, we consider a scenario for downlink LTE system for a mobile speed equal to 350 Km/h. Accordingly, we take into account the impulse noise with $p = 0.2$ which was added to the reference signals with different rates of SIR and it ranged from -10 to 10 dB.

International Journal of Wireless & Mobile Networks (IJWMN) Vol. 3, No. 4, August 2011

Table 2. Parameters of simulations [15],[16] and [17].

| Parameters | Specifications |
|---|---|
| OFDM system | LTE/Downlink |
| Constellation | 16-QAM |
| Mobile Speed (Km/h) | 350 |
| $T_s$ (μs) | 72 |
| $f_c$ (GHz) | 2.15 |
| $\delta f$ (KHz) | 15 |
| $B$ (MHz) | 5 |
| Size of DFT/IDFT | 512 |
| Number of paths | 9 |

Figure (1) presents the variations in time and in frequency of the channel frequency response for the considered scenario.

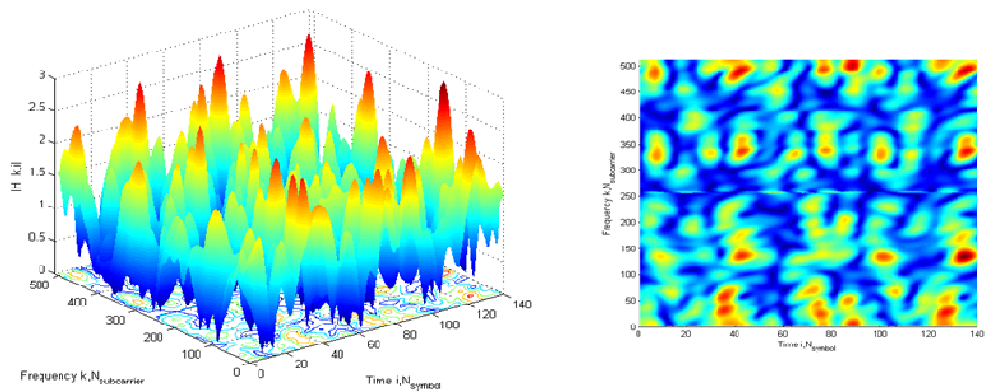

Figure 1. Variations in time and in frequency of the channel frequency response for mobile speed =350 Km/h.

Figure (2) shows the performance of the LS and nonlinear complex LS-SVM algorithms in the presence of additive Gaussian noise as a function of SNR for SIR=0 and -10 dB respectively. A poor performance is noticeably exhibited by LS and better performance is observed with nonlinear complex LS-SVM.

International Journal of Wireless & Mobile Networks (IJWMN) Vol. 3, No. 4, August 2011

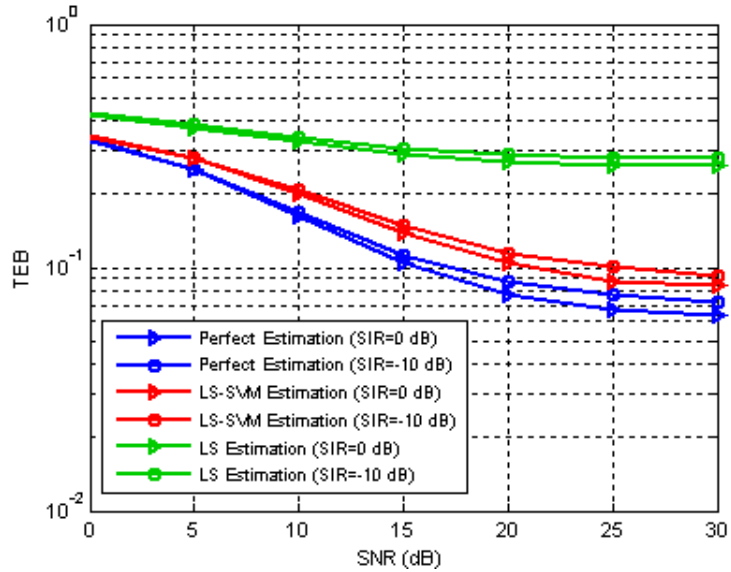

Figure 2. BER as a function of SNR for a mobile speed at 350 Km/h for SIR=0 and -10 dB with p=0.2.

Figure (3) presents a comparison between LS and nonlinear complex LS-SVM in the presence of additive impulse noise for SNR=20 dB. The comparison of these methods reveals that nonlinear complex LS-SVM outperform LS estimator in high mobility conditions especially for high SNR as confirmed by figure (4) for SNR=30 dB.

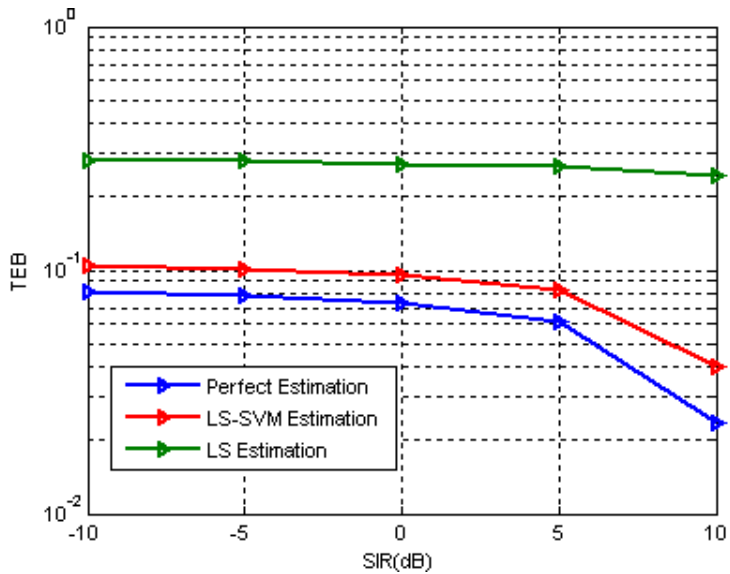

Figure 3. BER as a function of SIR for a mobile speed at 350 Km/h for SNR=20 dB with p=0.2.

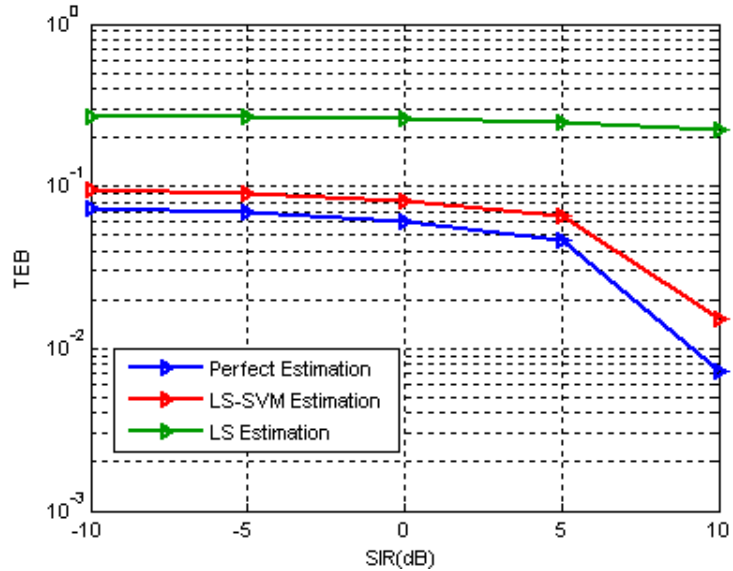

Figure 4.  BER as a function of SIR for a mobile speed at 350 Km/h for SNR=30 dB with p=0.2.

## 5. CONCLUSION

In this contribution, we have presented a new nonlinear complex LS-SVM based channel estimation technique for a downlink LTE system under high mobility conditions in the presence of non-Gaussian impulse noise interfering with OFDM reference symbols.

The proposed method is based on learning process that uses training sequence to estimate the channel variations. Our formulation is based on nonlinear complex LS-SVM specifically developed for pilot-based OFDM systems. Simulations have confirmed the capabilities of the proposed nonlinear complex LS-SVM in the presence of Gaussian and impulse noise interfering with the pilot symbols for a high mobile speed when compared to LS standard method. The proposal takes into account the temporal-spectral relationship of the OFDM signal for highly selective channels. The Gram matrix using RBF kernel lead to a significant benefit for OFDM communications especially in those scenarios in which impulse noise and deep fading are presents.